%% file: iclr2024_conference.tex
\documentclass{article} 
\usepackage{iclr2024_conference,times}

\input{math_commands.tex}

\usepackage{hyperref}
\usepackage{url}

\usepackage{graphicx}
\usepackage{amsmath}
\usepackage{amssymb}
\usepackage{booktabs}
\usepackage{dsfont}
\usepackage{pifont}
\usepackage{soul}
\usepackage{colortbl}
\usepackage{subcaption}
\usepackage{multicol}
\usepackage{bbm}
\usepackage{wrapfig,lipsum,booktabs}

\usepackage{cuted}
\usepackage{capt-of}

\usepackage[capitalize]{cleveref}
\crefname{section}{Sec.}{Secs.}
\Crefname{section}{Section}{Sections}
\Crefname{table}{Table}{Tables}
\crefname{table}{Tab.}{Tabs.}

\title{CrIBo \includegraphics[width=5.5mm]{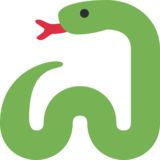}: Self-Supervised Learning via Cross-Image Object-Level Bootstrapping}

\author{
\centerline{Tim Lebailly$^{1}$\thanks{denotes equal contribution.} \setcounter{footnote}{0} \hspace{1cm} Thomas Stegm\"uller$^{2\hspace{0.035cm}*}$ \hspace{1cm} Behzad Bozorgtabar$^{2,3}$}\\
\centerline{\textbf{Jean-Philippe Thiran}$^{2,3}$ \hspace{1cm} \textbf{Tinne Tuytelaars}$^{1}$} \\
\centerline{$^{1}$KU Leuven \hspace{1cm} $^{2}$EPFL \hspace{1cm} $^{3}$CHUV} \\ 
{\small $^{1}$\texttt{\{firstname\}.\{lastname\}@esat.kuleuven.be} \hfill $^{2}$\texttt{\{firstname\}.\{lastname\}@epfl.ch}}
}

\iclrfinalcopy 
\begin{document}

\input{chapters/definitions}
\maketitle

\begin{abstract}
\input{chapters/abstract}

\end{abstract}

\input{chapters/introduction}

\input{chapters/related_works}
\input{chapters/method}
\input{chapters/experiments}

\input{chapters/conclusion}

\clearpage

\noindent
{\bf Acknowledgement.}
\label{par:Acknowledgement}
This project is funded by the European Research Council (ERC) under the European Union’s Horizon 2020 research and innovation program (Grant Agreement No. 101021347). This work is also partially funded by the Personalized Health and Related Technologies (PHRT), grant number 2021/344; as well as the Fonds Wetenschappelijk Onderzoek (FWO), project G0A4720N. We acknowledge both EuroCC Belgium and the Swiss National Supercomputing Centre (project ID 606) for awarding this project access to the LUMI supercomputer.

\bibliography{iclr2024_conference}
\bibliographystyle{iclr2024_conference}

\input{chapters/appendix}
\end{document}

%% file: math_commands.tex
\usepackage{amsmath,amsfonts,bm}

\def\eqref#1{equation~\ref{#1}}

\def\1{\bm{1}}

\DeclareMathAlphabet{\mathsfit}{\encodingdefault}{\sfdefault}{m}{sl}
\SetMathAlphabet{\mathsfit}{bold}{\encodingdefault}{\sfdefault}{bx}{n}

\DeclareMathOperator*{\argmax}{arg\,max}
\DeclareMathOperator*{\argmin}{arg\,min}

%% file: chapters/definitions.tex
\newcommand{\TODO}[1]{{\color{red} {\bf TODO:} #1}}

\newcommand{\Tim}[1]{\textcolor{magenta}{#1}}
\newcommand{\Thomas}[1]{\textcolor{blue}{#1}}
\newcommand{\Behzad}[1]{\textcolor{red}{#1}}
\newcommand{\Tinne}[1]{\textcolor{green}{#1}}
\newcommand{\Vestige}[1]{\textcolor{gray}{#1}}

\newcommand{\aug}{\tilde{\boldsymbol{x}}}
\newcommand{\im}{\boldsymbol{x}}
\newcommand{\glob}{\bar{\boldsymbol{z}}}
\newcommand{\z}{\boldsymbol{z}}

\newcommand{\cent}{\boldsymbol{c}}
\newcommand{\q}{\boldsymbol{q}}
\newcommand\norm[1]{\left\lVert#1\right\rVert}

\newcommand{\NN}{\texttt{nn}}
\newcommand{\jump}{\texttt{jump}}
\newcommand{\cyclecond}{\mathcal{O}}

\newcommand{\dense}{\boldsymbol{z}}
\newcommand{\pos}{\mathbf{E}}
\newcommand{\objects}{\mathbf{O}}
\newcommand{\centroids}{\mathbf{C}}
\newcommand{\costsem}{\mathbf{T}^{\text{(sem)}}}
\newcommand{\costpos}{\mathbf{T}^{\text{(pos)}}}
\newcommand{\costtot}{\mathbf{T}^{\text{(tot)}}}
\newcommand{\loc}{\boldsymbol{z}_{t,k}}
\newcommand{\p}{\boldsymbol{p}}
\newcommand{\PP}{\mathbf{P}}
\newcommand{\Q}{\mathbf{Q}}
\newcommand{\Y}{\mathbf{Y}}
\newcommand{\mask}{\mathbf{M}}
\newcommand{\tmarg}{\mathbf{r}}
\newcommand{\cmarg}{\mathbf{c}}
\newcommand{\cls}{\texttt{[CLS]}}
\newcommand{\model}{g}
\newcommand{\backbone}{f}
\newcommand{\head}{h}

\newcommand{\mname}{\text{CrIBo}}
\newcommand{\slotcon}{\text{SlotCon}}
\newcommand{\vit}{\text{ViT}}
\newcommand{\croc}{\text{CrOC}}
\newcommand{\dino}{\text{DINO}}
\newcommand{\mae}{\text{MAE}}
\newcommand{\ibot}{\text{iBOT}}
\newcommand{\beit}{\text{BEiT}}
\newcommand{\loca}{\text{LOCA}}
\newcommand{\densecl}{\text{DenseCL}}
\newcommand{\resnet}{\text{ResNet}}
\newcommand{\soco}{\text{SoCo}}
\newcommand{\pixpro}{\text{PixPro}}
\newcommand{\resim}{\text{ReSim}}
\newcommand{\vicregl}{\text{VICRegL}}
\newcommand{\orl}{\text{ORL}}
\newcommand{\byol}{\text{BYOL}}
\newcommand{\detco}{\text{DetCo}}
\newcommand{\odin}{\text{ODIN}}
\newcommand{\hb}{\text{Hummingbird}}
\newcommand{\timet}{\text{TimeT}}
\newcommand{\cpsq}{\text{CP}^{2}}
\newcommand{\overbar}[1]{\mkern 1.5mu\overline{\mkern-1.5mu#1\mkern-1.5mu}\mkern 1.5mu}
\newcommand{\supp}{\textbf{Supplementary Material}}

\definecolor{light_cyan}{HTML}{c6effc}
\definecolor{light_green}{HTML}{72B54D}
\sethlcolor{light_cyan}

\newcommand{\ck}{\textcolor{green!80!black}{\ding{51}}}
\newcommand{\xk}{\textcolor{red}{\ding{55}}}

\newcommand{\etal}{\textit{et al}., }
\newcommand{\ie}{\textit{i}.\textit{e}., }
\newcommand{\eg}{\textit{e}.\textit{g}., }

%% file: chapters/abstract.tex
Leveraging nearest neighbor retrieval for self-supervised representation learning has proven beneficial with object-centric images. However, this approach faces limitations when applied to scene-centric datasets, where multiple objects within an image are only implicitly captured in the global representation. Such global bootstrapping can lead to undesirable entanglement of object representations. Furthermore, even object-centric datasets stand to benefit from a finer-grained bootstrapping approach. In response to these challenges, we introduce a novel \textbf{Cr}oss-\textbf{I}mage Object-Level \textbf{Bo}otstrapping method tailored to enhance dense visual representation learning. By employing object-level nearest neighbor bootstrapping throughout the training, CrIBo emerges as a notably strong and adequate candidate for in-context learning, leveraging nearest neighbor retrieval at test time. CrIBo shows state-of-the-art performance on the latter task while being highly competitive in more standard downstream segmentation tasks. Our code and pretrained models are publicly available at \texttt{\color{magenta}{https://github.com/tileb1/CrIBo}}.

%% file: chapters/introduction.tex
\begin{figure}[b]
    \centering
    \includegraphics[width=\textwidth]{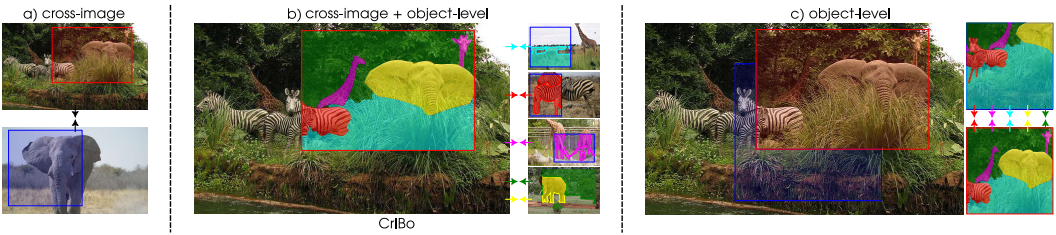}
    \caption{\textbf{Positioning of CriBO in the landscape of self-supervised learning.} \textbf{a)} Illustration of the cross-image self-supervision concept. \textbf{c)} Depiction of object-level self-supervision. \textbf{b)} CrIBo benefits from both learning paradigms. 
    }
    \label{fig:attrape_oeuil}
\end{figure}
\section{Introduction}
\label{sec:introduction}
Over the past few years, the field of artificial intelligence has experienced significant advancements, primarily driven by the democratization of deep learning techniques \citep{krizhevsky2012imagenet}. Self-supervised learning (SSL) stands out as one of the major factors contributing to the success of deep learning. Indeed, SSL has opened the door to the training of large foundation models pretrained on massive amounts of uncurated data. In natural language processing (NLP), this large-scale task-agnostic pretraining has been the key to unlocking general-purpose features that perform well on various downstream tasks. Conversely, in computer vision, task-specific models and finetuning remain the \textit{de facto} choice, in part due to the lack of consensus around the pretraining pretext task, such as is the case in NLP with masked language modeling (MLM). \par

Nonetheless, \cite{balavzevic2023towards} showed that under the auspices of their contextual pretraining, dense nearest neighbor retrieval performed comparably with task-specific models while being significantly more efficient. The contextualization mechanism ensures the alignment of the pretraining phase with dense nearest neighbor evaluation by leveraging within and across image attention. Nevertheless, this mechanism only partially fulfills the alignment objective as \textit{i)} the cross-image attention mechanism occurs from \textbf{local to global} features and \textit{ii)} consistency is only encouraged between contextualized \textbf{image-level} representations of views from the \textbf{same image}. To mitigate these limitations, we propose to \textbf{explicitly} enforce cross-image consistency between object-level representations as shown in \Cref{fig:attrape_oeuil}. \par
Most recent self-supervised learning algorithms, \eg \cite{chen2020simple,he2020momentum,grill2020bootstrap,caron2021emerging}, revolve around the objective of learning image-level representations invariant to semantic-preserving data augmentations. $\mname$ departs from this paradigm in two ways. First, we encourage consistency at the granularity of the object in the images, and second, this consistency is enforced between objects from different images. In doing so, our pretraining aligns well with the objective of obtaining general-purpose representations tailored for dense nearest neighbor retrieval. Furthermore, by operating at the object level, $\mname$ elegantly mitigates the pitfall of contextual bias \cite{singh2020don}, and is therefore compatible with \textit{scene-centric} images which constitute the bulk of web-scale datasets. \par

Our contributions are as follows. \textbf{1)} To the best of our knowledge, $\mname$ is the first end-to-end online SSL approach that explicitly enforces cross-image consistency at the object/object-part level. \textbf{2)} The resulting representations excel in \textit{in-context scene understanding} tasks \citep{balavzevic2023towards} without compromising the performance on standard segmentation benchmarks. \textbf{3)} Moreover, $\mname$ is compatible with scene-centric datasets, addressing a gap in the existing cross-image SSL literature.

%% file: chapters/related_works.tex
\section{Related works}
\label{sec:related_works}
\noindent

{\bf Image-level self-supervision.}
\label{par:global_features}
Seminal work SimCLR \citep{chen2020simple} has played a pivotal role in promoting the widespread adoption of \textit{cross-view consistency} as a fundamental principle for visual representation learning. However, such pretext task admits trivial solutions and contemporary works have yet to agree on the ideal way to prevent them. One effective and intuitive approach to address this issue involves the use of negative samples \citep{chen2020simple,hjelm2018learning}, which effectively mitigates degeneracy but requires the use of large batchsizes. This can be alleviated with a memory bank \citep{he2020momentum}. Self-distillation methods, on the other hand, do not use explicit negatives. They avoid trivial solutions by using asymmetry in the form of \eg an additional predictor \citep{simsiam,byol} on one branch, using stop-gradients \citep{simsiam,caron2021emerging,esvit,byol}, a momentum encoder \citep{caron2021emerging,byol}, \textit{etc}. On the other end of the spectrum, clustering-based methods avoid trivial solutions by regularizing the assignment of the samples across a set of clusters \citep{asano2019self,caron2018deep,caron2020unsupervised,caron2021emerging,Zhuang_2019_ICCV}.

\noindent
{\bf Localized self-supervision.}
\label{par:local_features}
A finer-grained self-supervision can be obtained by leveraging the inherent spatial structure of images and by enforcing cross-view consistency at a localized level. This ensures that the resulting features are better aligned with dense downstream tasks. These methods can be broadly categorized based on the level of granularity at which they enforce similarity. One line of work involves enforcing similarity at the feature level, known as the local-level approach, where individual features (or local-representations) are directly contrasted \citep{wang2021dense,liu2020self,o2020unsupervised,xie2021propagate,lebailly2022global,bardes2022vicregl}. Another approach operates at a coarser level, promoting similarity between semantically coherent groups of features, referred to as the object-level approach \citep{cho2021picie,henaff2021efficient,henaff2022object,xie2021unsupervised,seitzer2022bridging,Stegmuller_2023_CVPR,wen2022self}.
\par

Alternatively, there have been propositions of dense fine-tuning strategies \citep{hamilton2022unsupervised,ziegler2022self,yun2022patch,wang2022cp2,zadaianchuk2022unsupervised}. These techniques are designed to enhance models initially trained with an image-level objective \citep{caron2021emerging} with localized self-supervision. However, the former cannot be used as stand-alone pretraining methods. Another way of using the inherent spatial structure is by leveraging patch location prediction \citep{caron2022location} or by asking the model to reconstruct the input image from a corrupted/masked version thereof \citep{he2022masked}.

\noindent
{\bf Cross-image self-supervision.}
\label{par:cross_image}
Moving beyond the conventional cross-view consistency inherent in numerous self-supervised methods, image-level bootstrapping harnesses nearest neighbors in the latent space for self-labeling and metric learning \citep{dwibedi_little_2021,Koohpayegani_2021_ICCV,DBLP:journals/corr/abs-2102-10106,Xie2022delving}. Current bootstrapping methods for representation learning face two challenges. Firstly, the effectiveness of using distance metrics in the latent space as a proxy for semantic closeness is contingent on the quality of the learned representations, especially during the initial stage of pretraining when the encoder is randomly initialized. To address this limitation, adaptive bootstrapping methods have been proposed, aiming to avoid systematic nearest neighbor bootstrapping \citep{lebailly2023adaptive}. Secondly, in the context of scene-centric images, image-level bootstrapping reveals a tendency to
entangle object representations, which can have detrimental effects on dense downstream tasks. These issues motivate the introduction of the object-level cross-image self-supervised learning paradigm. To that end, $\orl$~\citep{xie2021unsupervised} introduced a three-stage procedure to identify pairs of region-of-interests (RoIs) and encourage cross-image consistency between cropped objects at the image-level.

Recently, \cite{balavzevic2023towards} proposed to leverage attention within and across images. The latter contextualizes the local representations of an image with the global ones from other images to adapt for in-context learning of various scene understanding tasks at test time.

%% file: chapters/method.tex
\section{Method}
\label{sec:method}

\subsection{Preliminaries} 
\label{ssec:preliminaries}
Before diving into the method, we briefly introduce the terminology used in this paper.
\paragraph{Dense representation.}
\phantomsection
\label{par:dense_representation}
A representation that explicitly preserves the spatial dependencies of the input image is referred to as a \textit{dense representation}. We denote such representations by $\dense \in \mathbb{R}^{H \times W \times d}$ or $\in \mathbb{R}^{HW \times d}$ where $H$ and $W$ are the height and width of the image or a downscaled version thereof and $d$ refers to the dimension of the latent space. The dense representation produced by a convolutional neural network is a stack of output feature maps, whereas a vision transformer yields an ordered sequence of tokens.

\paragraph{Local representation.}
\phantomsection
\label{par:local_representation}
 Given a spatial location $n$ out of the $N=HW$ possible ones, the associated slice of the dense representation $\dense^{n} \in \mathbb{R}^{d}$ with $n \in \{1, 2, \cdots, N\}$ is referred to as a \textit{local representation}.

\paragraph{Object representation.}
\phantomsection
\label{par:object_representation}
The \textit{object representation} $\cent^k \in \mathbb{R}^{d}$ of the $k^\text{th}$ object within an image is derived by aggregating the local representations associated with the corresponding object. We use the term \textit{object} loosely to refer to a semantically coherent region within the image.

\paragraph{Global representation.}
\phantomsection
\label{par:global_representation}
The \textit{global representation}, denoted as $\glob \in \mathbb{R}^{d}$, is a global aggregation over the spatial dimensions of the dense representation. As an example, this corresponds to the $\texttt{[CLS]}$ token in a $\vit$~\citep{dosovitskiy2020image} or the result of a global average pooling layer in case of a CNN.
\paragraph{Bootstrapping}
\phantomsection
\label{par:bootstrapping}
Here, the term \textit{bootstrapping}\footnote{The term bootstrapping is employed in the colloquial sense, as opposed to the statistical sense.} refers to the utilization of the model's current knowledge to enhance its performance further.  In the scope of this study, this materializes as leveraging distances in the latent space to find object-level nearest neighbors that can be used to form positive pairs for self-distillation.

\subsection{Object-level Cross-image Bootstrapping (CrIBo)}
\label{ssec:object_level_bootstrapping}
In~\Cref{sec:introduction}, we discuss the need for generalist models capable of performing various downstream tasks without requiring any finetuning. A recent study~\citep{balavzevic2023towards} demonstrated that this could be achieved by leveraging nearest neighbor retrieval at test time and contextualization throughout the training phase. Aligning with this aim, we propose further reducing the gap between the training and test phases by explicitly enforcing consistency between object-level nearest neighbors. The existing SSL studies exploiting nearest neighbors are built solely around global representations and are rooted in the assumption of datasets being object-centric. 
Hence, a simplistic extension of such frameworks to our proposed scenario is impracticable.
To accomplish our goal, the following steps are required:
\begin{enumerate}
    \item Identify semantically coherent regions within the image to create object-level representations (\cref{subsec:semanticallycoherentimageregions}).
    \item Match pairs of object-level representations across images (\cref{subsec:cross-image-matchings}).
    \item Apply cross-image consistency between pairs of matching object-level representations (\cref{subsec:resulting_ssl}).
\end{enumerate}
A high-level overview of our proposed method $\mname$ can be found in \Cref{fig:main_fig}.

\subsubsection{Semantically coherent image regions}
\label{subsec:semanticallycoherentimageregions}
The first step towards cross-image object-level bootstrapping is to identify semantically coherent image regions within the images. We argue that it is beneficial to use pairs of augmented views for each input image and to have a cross-view correspondence between the objects in both views. Indeed, image-level SSL methods yield significantly worsened performance when training without photometric augmentations \citep{dwibedi_little_2021}. This is observed when positive pairs originate from the same image and to a lesser extent (but still significant) when they do not. This drop in performance highlights the need to shed away low-level similarities between positive pairs of images with data augmentations. To prevent this from occurring with nearest neighbor (NN) positives, \cite{dwibedi_little_2021} refrain from enforcing consistency between a retrieved NN and its query. At the image level, this can be easily implemented by leveraging the duplicity of the views. More specifically, given two augmented views $\aug_1$ and $\aug_2$ as well as their corresponding global-representation $\glob_1$ and $\glob_2$, a similarity constraint is enforced between $\glob_1$ and $\texttt{nn}(\glob_2)$ where $\texttt{nn}(\cdot)$ denotes the nearest neighbor operator. Conversely, at the object level, this requires a cross-view correspondence between object representations from both views. To that end, we rely on an online joint-space clustering algorithm from~\cite{Stegmuller_2023_CVPR}.
\par

\paragraph{Joint-space clustering.}
\phantomsection
\label{par:joint_space_clustering}
\input{figures/bootstrapping_sketch}
The clustering algorithm takes as input the dense representations $\z_1 \in \mathbb{R}^{N \times d}$ and $\z_2 \in \mathbb{R}^{N \times d}$ associated with the two augmented views and concatenates them along the token axis to obtain the dense representation of the joint-view, $\z_{\text{cat}} \in \mathbb{R}^{2N \times d}$. The $2N$ tokens are subsequently partitioned into $K$ clusters to produce semantically coherent and spatially compact clusters. A hyperparameter $\lambda_{\text{pos}}$ modulates the importance of the spatiality against that of the semantic (see~\cref{tab:ablation_hp}). The resulting assignment is encoded in a matrix $\Q^{*} \in \mathbb{R}^{2N \times K}$, which can be split in two to obtain the view-wise assignments $\Q^{*}_1, \Q^{*}_2 \in \mathbb{R}^{N \times K}$. Here $\Q^{*}_i(n,k)$ denotes a soft-assignment of token $n$ to object-cluster $k$ in view $i$. These can in turn be used to compute the $K$ object representations $\cent_i \in \mathbb{R}^{d}$ in each view $i$ as follows:
\begin{equation}
    \cent_i^k = \sum_{n \in [N]} \Q^{*}_i(n,k) \cdot \z_i^n,\quad \quad i \in \{1,2\}
    \label{eq:vanilla_object_representation}
\end{equation}
Note that the columns of $\Q^{*}_i$ undergo $l_1$ normalization, rendering the object representations as affine combinations of local representations. It can be observed that the cross-view correspondence is provided for free as the object representation $\cent_1^k$ from the first view is matched to $\cent_2^k$ in the second one as a consequence of the joint-space clustering. In practice, it can be the case that some objects are exclusive to one of the two views \ie $\sum_{n \in [N]} \Q^{*}_i(n,k) = 0$. In that case, the object $k$ is simply discarded from the view where it is present. In~\Cref{par:cycle-consistent-matching}, we show that such cross-view correspondence can be harnessed to improve the training stability through a bootstrapping criterion.

\subsubsection{Cross-image object matchings}
\label{subsec:cross-image-matchings}
In~\Cref{subsec:semanticallycoherentimageregions}, we detail the methodology for identifying objects within the views of a given image and establish the strategy for their alignment from one view to another. Subsequent to this step is the critical phase of aligning such objects across images. For reasons that will be delineated in the next paragraph, we rely on a pair of memory banks $\mathcal{M}_1$ and $\mathcal{M}_2$, which are populated with object representations extracted from the first and second views during previous iterations. As such, the memory banks $\mathcal{M}_{i}$ can be queried with an object representation $\cent_i^k$ to retrieve its nearest neighbor:
\begin{equation}
\label{eq:nn_operator}
    \NN(\cent_i^k, \mathcal{M}_i) \triangleq \argmax_{\cent \in \mathcal{M}_i} \frac{{\cent}^\top \cent_i^k}{\norm{\cent}_2 \norm{\cent_i^k}_2}
\end{equation}
Here, we consider $\NN(\cent^k_1, \mathcal{M}_{1})$ and $\cent_1^k$ as cross-image object-level matches. A visualization of cross-image object-level matches can be found in \Cref{ssec:imagenetmatchings} (\cref{fig:matching_coco} and \cref{fig:matching_in}).

\paragraph{Cycle-consistent matchings.}
\phantomsection
\label{par:cycle-consistent-matching}
\begin{figure}
    \centering
    \includegraphics[width=\textwidth]{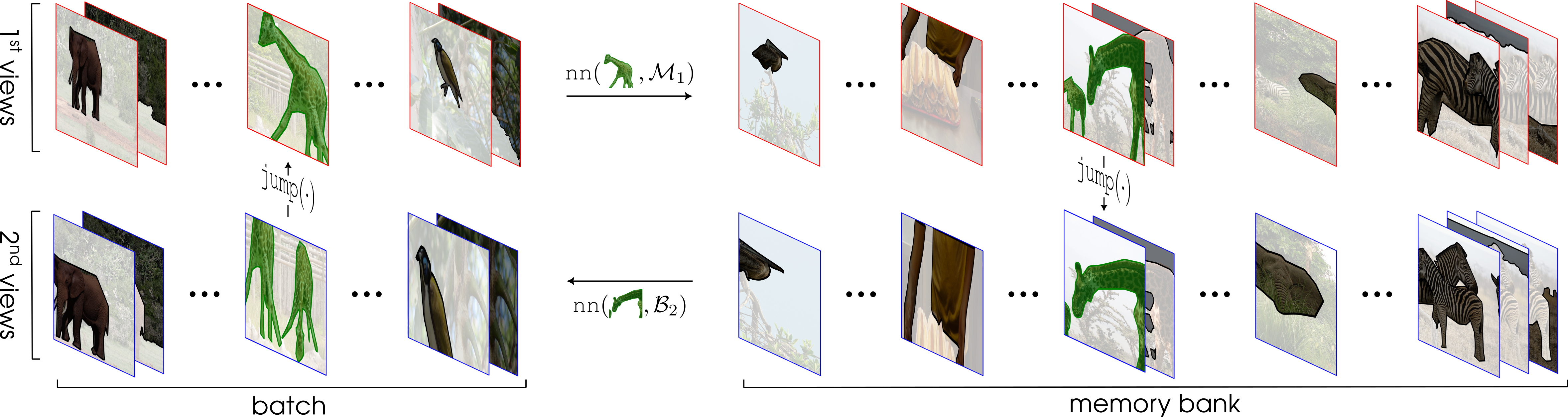}
    \caption{\textbf{Illustration of cycle-consistent matchings}. Such matchings are invariant to data augmentations and reciprocal, loosely speaking.}
    \label{fig:cycle_consistency}
\end{figure}
At the beginning of the pretraining phase, when the encoder is randomly initialized, relying on~\Cref{eq:nn_operator} to identify cross-image object-level matches is not well grounded. Indeed, this straightforward approach assumes that the representations of similar objects are close in the embedding space, an assumption that does not hold at the initialization. Therefore, enforcing consistency between such pairs of semantically distinct objects should be avoided. To that end, we design a \textit{semantic closeness} condition, which is only fulfilled when the matches are established based on high-level features, thus enabling the discarding of invalid positive pairs that do not share a semantic similarity.
 \par

At a high-level, we consider a pair of objects as a valid positive pair if they are the nearest neighbors of each other. In practice, the criterion is implemented as follows. Starting from an object representation $\cent_1^k$ in the batch of first views, its nearest neighbor $\NN(\cent_1^k, \mathcal{M}_1)$ in the queue is found using~\Cref{eq:nn_operator}. To ensure that the criterion is invariant to data augmentation, we then switch to the queue of second views via the $\jump(\cdot)$ operator:
\begin{equation}
    \jump(\cent_1^k) \triangleq \cent_2^k, \quad \quad \jump(\cent_2^k) \triangleq \cent_1^k
\end{equation}
From this step, we obtain $\jump(\NN(\cent_1^k, \mathcal{M}_1))$ and retrieve its NN in the batch of second views $\NN(\jump(\NN(\cent_1^k, \mathcal{M}_1)), \mathcal{B}_{2})$. The final step consists of switching back to the batch of first views and verifying if the resulting object is the same as the one we started from. More formally:
\begin{equation}
  \cyclecond(\cent_i^k) \triangleq
    \begin{cases}
      true & \text{if} \quad \cent_i^k = \jump(\NN(\jump(\NN(\cent_i^k, \mathcal{M}_i)), \mathcal{B}_j))\\
      false & \text{otherwise}
    \end{cases}       
    \quad i, j \in \{1, 2\}, i \neq j
\end{equation}
A depiction of cycle-consistent matchings is provided in \Cref{fig:cycle_consistency}.

\subsubsection{Self-supervised training objectives}
\label{subsec:resulting_ssl}
At this point, we have all the necessary components to apply three distinct types of self-supervision: cross-view object-level, cross-image object-level, and global cross-image. All levels of self-supervision in our method CrIBo are based on the self-distillation mechanism proposed by \cite{caron2021emerging}, which employs a student-teacher pair of siamese networks $f_s$ and $f_t$, respectively. Given two augmented views $\aug_1$ and $\aug_2$, global representations and object representations can be obtained for each view and each network within the siamese pair.
\paragraph{Cross-View Object-Level Self-Supervision.}
\phantomsection
\label{par:dense-cross-view-self-supervision}
The object representations from the teacher and student are fed to their respective head $h_s$ and $h_t$ to output probability mass functions whose sharpness can be modulated via temperature parameters ($\tau_{s}$ and $\tau_{t}$). For the teacher, this translates to:
\begin{align}
    \label{eq:projection}
    \begin{split}
    \p_{i,t}^k &= \underset{L}{\texttt{softmax}}\left(\head_{t}(\cent_{i,t}^k) / \tau_{t} \right),  \quad \quad i \in \{1,2\}\\
    \end{split}
\end{align}
The student's projections are obtained analogously. The cross-view object-level loss is expressed as the averaged cross-entropy over all pairs of object representations and making the loss symmetric w.r.t. the student/teacher:
\begin{equation}
    \label{eq:dense_cross_view_loss}
\mathcal{L}_{cv}^{o} = \frac{1}{K}\sum_{k \in [K]} \left( H(\p_{1,t}^k, \p_{2,s}^k) +  H(\p_{2,t}^k, \p_{1,s}^k) \right)
\end{equation}
where $H(\mathbf{a}, \mathbf{b}) = -\sum_{l=1}^{L} \mathbf{a}_{l} \log (\mathbf{b}_{l})$ denotes the cross-entropy.

\paragraph{Cross-Image Object-Level Self-Supervision.}
\phantomsection
\label{par:dense-cross-image-self-supervision}
Once all object representations in the batch have been retrieved, the cross-image object-level loss is enforced in a manner analogous to its cross-view counterpart. The nearest neighbors are first fed to the projection head as follows:
\begin{align}
    \label{eq:projectionnn}
    \begin{split}
    \tilde{\p}_{i,t}^{k} &= \underset{L}{\texttt{softmax}}\left(\head_{t}(\texttt{nn}(\cent_{i,t}^k)) / \tau_{t} \right),  \quad \quad i \in \{1,2\}\\
    \end{split}
\end{align}

Note that these projections are only defined for the teacher since the $\NN(\cdot)$ operator is not differentiable. The formulation of the cross-image object-level loss  aligns with that in~\Cref{eq:dense_cross_view_loss} up to the filtering of invalid positive pairs (see~\cref{par:cycle-consistent-matching}):
\begin{equation}
    \label{eq:dense_cross_image_loss}
    \mathcal{L}_{ci}^{o} = \frac{1}{Z_1}\sum_{k \in [K]} \mathbbm{1}_{\{\cyclecond(\cent_1^k)\}} H(\tilde{\p}_{1,t}^{k}, \p_{2,s}) +  \frac{1}{Z_2}\sum_{k \in [K]} \mathbbm{1}_{\{\cyclecond(\cent_2^k)\}} H(\tilde{\p}_{2,t}^{k}, \p_{1,s})
\end{equation}
where $\mathbbm{1}_{\{\cdot\}}$ denotes the indicator function and where $Z_1$ and $Z_2$ are normalization constants \ie $\sum_{k \in [K]} \mathbbm{1}_{\{\cyclecond(\cent_1^k)\}}$ and $\sum_{k \in [K]} \mathbbm{1}_{\{\cyclecond(\cent_2^k)\}}$, respectively. Empirical evidence on the utility of the cycle consistency can be found in \Cref{ssec:ablationstudy_cc}.

\paragraph{Global Cross-View Self-Supervision.}
\phantomsection
\label{par:global-cross-view-self-supervision}
Importantly, both the clustering and the bootstrapping steps are contingent on the quality of the representations. To avoid potential instabilities, we incorporate a global cross-view loss into our framework, offering meaningful self-supervision at any stage of the pretraining. This loss is analogous to the one from \Cref{eq:dense_cross_view_loss} except that it is applied to the global representation ($\z_{1,s}$, $\z_{2,s}$, $\z_{1,t}$ and $\z_{2,t}$), which are fed to a dedicated head $\overbar{h}$:
\begin{align}
    \label{eq:image_proj}
    \begin{split}
    \overbar{\p}_{i,t} &= \underset{\overbar{L}}{\texttt{softmax}} \left(\overbar{\head}_{t}(\glob_{i,t} / \overbar{\tau}_{t})\right), \quad \quad i \in \{1,2\} \\
    \end{split}
\end{align}
where $\overbar{\tau}_{s}$ and $\overbar{\tau}_{t}$ are the corresponding temperature parameters for the $\overbar{L}$-dimensional output distribution. The global cross-view loss is defined as:
\begin{equation}
    \label{eq:glob_loss}
    \mathcal{L}_{cv}^{g} = H(\overbar{\p}_{1,t}, \overbar{\p}_{2,s}) +  H(\overbar{\p}_{2,t}, \overbar{\p}_{1,s})
\end{equation}
where $H(\mathbf{a}, \mathbf{b}) = -\sum_{l=1}^{\overbar{L}} \mathbf{a}_{l} \log (\mathbf{b}_{l})$.

Finally, we obtain the overall training objective of $\mname$ as a sum of its individual components:
\begin{equation}
    \mathcal{L}_{\text{tot}} = \mathcal{L}_{cv}^{g} + \mathcal{L}_{cv}^{o} + \mathcal{L}_{ci}^{o}
    \label{eq:total_loss}
\end{equation}
An ablation study over the inclusion of the individual loss terms can be found in \Cref{tab:ablation_loss}. As in \cite{caron2021emerging}, the student's parameters are optimized to minimize the above loss whereas the parameters of the teacher are updated via an exponential moving average of the student's weights.

%% file: figures/bootstrapping_sketch.tex
\begin{figure}
    \centering
    \includegraphics[width=\textwidth]{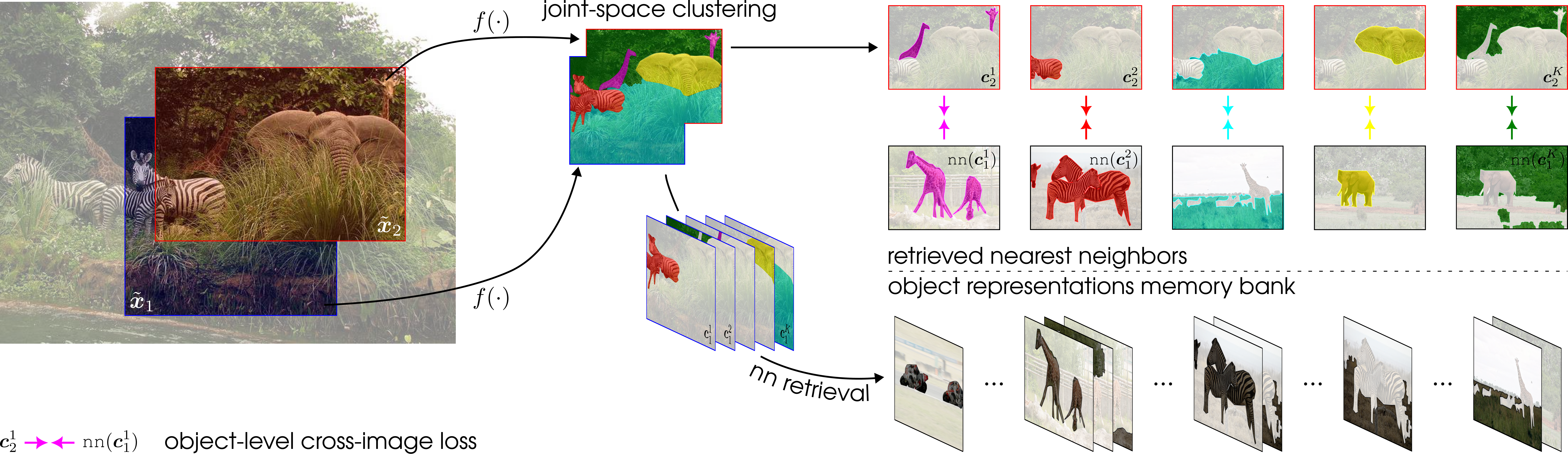}
    \caption{\textbf{High-level overview of cross-image object-level bootstrapping (CrIBo).} Given an encoder $f$ and pair of augmented views $\aug_1$ and $\aug_2$, object representations $\cent_i^k$ (depicted as colored object masks) from each view $i$ are computed. Using a memory bank, the nearest neighbors of each object representation $\cent_1^k$ from the first view are retrieved. A self-supervised consistency loss (depicted as colored arrows) is then enforced between $\cent_2^k$ and its corresponding retrieved neighbor from the other view $\NN(\cent_1^k)$.}
    \label{fig:main_fig}
\end{figure}

%% file: chapters/experiments.tex
\section{Experiments}
\label{sec:experiments}

\input{tables/hummingbird_new}

We first verify that the proposed pretraining method aligns well with the objective of in-context learning via nearest neighbor retrieval. We then show that in doing so, we do not compromise the performance on standard evaluations. General details on the experimental setup can be found in \Cref{ssec:experimental_setup}.

\subsection{Dense nearest neighbor retrieval}
\label{sssec:dense_nn_eval}
Inspired by~\cite{balavzevic2023towards}, we probe the quality of the learned features with a dense nearest neighbor (NN) classification task. The $k$-NN classifier is fitted on the local representations of a uniformly sub-sampled set of training images and evaluated on all the patches from the validation set of images. The sub-sampling factor is either 1, 8, 64 or 128. The only preprocessing applied to the images is a down-sampling to the closest resolution which is a multiple of the patch size. Following that step, patch-level labels are obtained via majority voting. We report the mIoU scores on Pascal VOC 2012~\citep{pascal-voc-2012} and ADE20K~\citep{zhou2017scene}. When the sub-sampling ratio is greater than 1, the reported mIoU scores result from averaging over 5 independent runs. The number of nearest neighbor patches is set to $k=50$. \par

In~\Cref{table:hummingbird}, we observe that using object-level nearest neighbor bootstrapping throughout training is an adequate pretext task for this evaluation, which is not a surprise due to the inherent similarities between the training and test phases. Overall, $\mname$ outperforms all methods, but Hummingbird~\citep{balavzevic2023towards}, by a significant margin and particularly following the scene-centric pretraining. It is worth noting that the comparison with \cite{balavzevic2023towards} is not exactly \textit{apple-to-apple}, as \textit{i)} we do not use multiple augmentations per image and \textit{ii)} we refrain from balancing the classes in the training set. These differences in the implementation of the evaluation reflect the divergence in our objectives: we place a higher emphasis on the assessment of the intrinsic quality of the learned representations rather than the absolute value of the reported results.

\input{tables/frozen_linear_segmentation_1token}

\subsection{Segmentation with a linear head}
\label{sssec:linear_segmentation_eval}
The frozen local representations are fed to the linear decoder head from Segmenter~\citep{strudel2021segmenter} using the MMSegmentation~\citep{mmseg2020} implementation. Herein, spatial tokens undergo a linear projection to the class space, followed by bilinear up-sampling to match the input image dimensions, enabling the application of pixel-wise cross-entropy loss. Performance metrics, in terms of mIoU scores, are reported on the same 4 datasets (see~\Cref{par:app_linear_segmentation_protocol} for more details). \par
As reported in~\Cref{table:linear_segmentation_1token}, $\mname$ exhibits commendable performance in linear segmentation, following both scene-centric and object-centric pretrainings. Among the methods using a ViT-B/16, CrIBo strongly outperforms other baselines. In the case of ViT-S/16, CrIBo also outperforms all baselines but with a smaller margin, due mostly to TimeT \citep{salehi2023time} being a particularly strong baseline. It is noteworthy that as opposed to prevailing studies, \eg~\cite{Stegmuller_2023_CVPR,bardes2022vicregl}, we do not concatenate spatial tokens from multiple layers of the $\vit$. Indeed, this trick is typically used to compare on an equal footing with $\resnet$-50, which has a much larger output dimension. As we only use and compare with $\vit$s, we opt for evaluations incorporating as minimal parameters as possible, aspiring to present a more accurate reflection of the features' intrinsic quality.

\input{tables/finetune_segmenter}
\subsection{End-to-end finetuning with Segmenter}
\label{sssec:segmenter_segmentation_eval}
In a real-world scenario, self-supervised pretraining often precedes end-to-end finetuning on the target dataset with an application-specific decoder head. The ability of the different pretraining paradigms to cope with that setting is assessed with an end-to-end finetuning of the backbone topped with the transformer-based decoder from Segmenter~\citep{strudel2021segmenter}. Here, the backbone's spatial features are fed to a transformer encoder along with $K$ learnable class tokens. The resulting class and spatial tokens are projected onto one another to obtain patch-level predictions, subsequently up-sampled to the input image's size to enforce the pixel-wise cross-entropy loss. We report the mIoU scores achieved on the same 4 datasets. \par 

The insights gained from the preceding experiments (\cref{table:hummingbird,table:linear_segmentation_1token}) underscore that models pretrained with $\mname$ have excellent generalists properties, \ie allowing them to perform various tasks without requiring finetuning. The results depicted in~\Cref{table:segmenter_segmentation} indicate that, when required, these models can also evolve into remarkable specialists. All that aside, after fine-tuning, all models tend to perform comparably, highlighting that the finetuning may be a suboptimal regime to compare SSL pretraining methods.

\input{tables/ablation_hp}

\subsection{Ablations}
\label{ssec:ablationstudy_hyperparam}

We report in~\Cref{tab:ablation_hp} the results of a grid search over various hyperparameters performed on COCO. Interestingly, we observe that $\mname$ performs the best when operating in the overclustering regime, that is, with a $K$ much larger than the number of objects present in the images, but only when bootstrapping is used. We postulate that this is due to the fact that overclustering offers the possibility to find better matches. To illustrate, consider the image in~\Cref{fig:main_fig} depicting a zebra, a giraffe, and an elephant. With $K=1$, this would necessitate identifying another image displaying the identical trio of animals; which is rare. With a larger value of $K$, the animals are isolated, expanding the number of good candidates. In the overclustering regime, the clusters represent object parts, which further facilitates the matchings. Similarly, a larger queue also amplifies the number of suitable candidates,  a trend reflected in the results until reaching a saturation point at a queue size of $25\text{k}$. More ablations are available in \Cref{sec:app_additional_experiments}.

%% file: tables/hummingbird_new.tex
\begin{table}[t]
\centering
\caption{
\textbf{Dense nearest neighbor retrieval}. The quality of the learned spatial features is probed with a k-NN classifier and using different ratios of training data. The depicted mIoU scores are derived from the validation sets of two scene-centric datasets. $\dagger$ refers to our own reproduction from official GitHub repositories. If not specified, publicly available checkpoints are used. $\star$ denotes results taken from \cite{balavzevic2023towards}.}
\vspace{0.2cm}
\footnotesize
\resizebox{1\textwidth}{!}{
\begin{tabular}{l c c c c c c c c c c c c c c c c c}
 &&&&& \multicolumn{4}{c}{ADE20K} && \multicolumn{4}{c}{Pascal VOC}  \\
\cmidrule{7-10}  \cmidrule{12-15}
Method & Backbone & Params & Dataset & Epochs && 1/128 & 1/64 & 1/8 & 1/1 && 1/128 & 1/64 & 1/8 & 1/1 \\
\midrule
\textit{Scene-centric} \\
$\slotcon$ \citep{wen2022self} & $\resnet$-50 & 25M & COCO & 800 && 9.9 & 11.8 & 17.3 & 22.1 && 37.3 & 42.8 & \textbf{52.9} & 57.2 \\
$\orl$ \citep{xie2021unsupervised} & $\resnet$-50 & 25M & COCO & 800 &&9.1 & 10.3 & 13.6 & 16.6 && 30.2 & 33.3 & 42.0 & 45.6\\

$\dino^{\dagger}$ \citep{caron2021emerging} & $\vit$-S/16 & 21M & COCO & 300 && 6.1 & 6.9 & 9.7 & 13.0 && 16.2 & 18.4 & 25.5 & 31.9  \\
$\mae^{\dagger}$ \citep{he2022masked} & $\vit$-S/16 & 21M & COCO & 300 && 3.7 & 4.1 & 5.4 & 6.8 && 8.5 & 9.3 & 12.2 & 15.9 \\
$\croc$ \citep{Stegmuller_2023_CVPR} & $\vit$-S/16 & 21M & COCO & 300 && 7.6 & 9.0 & 13.1 & 18.0 && 27.1 & 31.4 & 40.3 & 47.1  \\
\rowcolor{light_cyan} $\mname$ & $\vit$-S/16 & 21M & COCO & 300 && \textbf{10.9} & \textbf{12.8} & \textbf{18.4} & \textbf{23.4} && \textbf{39.1} & \textbf{44.0} & 52.8 & \textbf{58.1}  \\

\arrayrulecolor{black!30}\midrule[0.5pt]
\textit{Object-centric} \\
$\slotcon$ \citep{wen2022self} & $\resnet$-50 & 25M & IN1K & 200 && 10.6 & 12.0 & 18.1 & 24.0 && 40.7 & 44.7 & 56.4 & 61.6 \\
$\dino$ \citep{caron2021emerging} & $\vit$-S/16 & 21M & IN1K & 800 && 9.4 & 10.6 & 14.6 & 18.4 && 24.5 & 28.7 & 38.7 & 46.1  \\
$\croc$ \citep{Stegmuller_2023_CVPR} & $\vit$-S/16 & 21M & IN1K & 300 && 7.8 & 9.6 & 15.2 & 20.6 && 30.7 & 37.7 & 54.8 & 64.2  \\
$\timet$ \citep{salehi2023time}  & $\vit$-S/16 & 21M & IN1K+YTVOS & 800+30 && 11.7 & 13.6 & 19.7 & 24.6 && 38.9 & 44.3 & 56.0 & 62.7  \\

\rowcolor{light_cyan} $\mname$ & $\vit$-S/16 & 21M & IN1K  & 800 && \textbf{13.7} & \textbf{16.5} & \textbf{23.2} & \textbf{28.3} && \textbf{52.7} & \textbf{59.3} & \textbf{69.3} & \textbf{73.2}  \\

\arrayrulecolor{black!30}\midrule[0.5pt]
$\dino$ \citep{caron2021emerging} & $\vit$-B/16 & 85M & IN1K & 400 && 11.1 & 12.6 & 17.6 & 22.0 && 29.2 & 34.7 & 47.2 & 54.9 \\
$\mae$ \citep{he2022masked}  & $\vit$-B/16 & 85M & IN1K & 1600 && 2.7 & 3.0 & 4.0 & 5.3 && 6.0 & 6.5 & 8.9 & 13.8  \\
$\loca^{\star}$ \citep{caron2022location}  & $\vit$-B/16 & 85M & IN1K & 600 && - & - & - & 18.5 && - & - & - & 57.5  \\
$\hb^{\star}$ \citep{balavzevic2023towards}  & $\vit$-B/16 & 85M & IN1K & 300 && 11.7 & 15.1 & - & 28.3 && \textbf{50.5} & 57.2 & - & 70.5  \\

\rowcolor{light_cyan} $\mname$ & $\vit$-B/16 & 85M & IN1K  & 400 && \textbf{13.2} & \textbf{16.5} & \textbf{23.6} & \textbf{30.0} && \textbf{50.5} & \textbf{60.3} & \textbf{70.8} & \textbf{74.9}  \\

\arrayrulecolor{black}\midrule[0.5pt]
\end{tabular}
\label{table:hummingbird}
}
\end{table}%

%% file: tables/frozen_linear_segmentation_1token.tex
\begin{table}[t]
\centering
\caption{
\textbf{Linear segmentation with frozen backbones.} The linear decoder from Segmenter~\cite{strudel2021segmenter} is trained on the frozen spatial features obtained with various self-supervised learning methods. We report the mIoU scores achieved on the validation sets of 4 different datasets. $\dagger$ refers to our own reproduction from official GitHub repositories. If not specified, publicly available checkpoints are used.}

\vspace{0.2cm}
\footnotesize
\resizebox{1\textwidth}{!}{
\begin{tabular}{l c c c c c c c c c c c c c c c}
Method & Backbone & Params & Dataset  & Epochs && Pascal Context && Pascal VOC && COCO-Stuff 164K && ADE20K \\
\midrule
\textit{Scene-centric} \\
$\dino^{\dagger}$ \citep{caron2021emerging} & $\vit$-S/16 & 21M & COCO  & 300 && 27.3 && 43.9 && 19.9  && 18.9  \\
$\mae^{\dagger}$ \citep{he2022masked} & $\vit$-S/16  & 21M& COCO  & 300  && 18.1 && 25.6 && 11.3 && 11.0  \\
$\croc$ \citep{Stegmuller_2023_CVPR} & $\vit$-S/16 & 21M & COCO  & 300 && 30.0  && 51.7  && 22.6  && 20.5   \\
\rowcolor{light_cyan} $\mname$ & $\vit$-S/16 & 21M  & COCO & 300 && 37.5 && 61.0 && 29.1 && 27.5 \\
\rowcolor{light_cyan} $\mname$ & $\vit$-S/8 & 21M  & COCO & 300 && \textbf{40.6} && \textbf{63.9} && \textbf{31.2} && \textbf{28.3} \\
\arrayrulecolor{black!30}\midrule[0.5pt]
\textit{Object-centric} \\
$\dino$ \citep{caron2021emerging} & $\vit$-S/16 & 21M& IN1K  & 800 && 32.4  && 49.0  && 24.1  && 23.4  \\
$\croc$ \citep{Stegmuller_2023_CVPR} & $\vit$-S/16 & 21M & IN1K  & 300 && 37.7  && 68.1  && 30.2  && 27.6   \\
$\timet$ \citep{salehi2023time} & $\vit$-S/16 & 21M& IN1K+YTVOS  & 800+30  && 40.6  && 67.9  && 32.0  && 29.9   \\
\rowcolor{light_cyan} $\mname$ & $\vit$-S/16 & 21M & IN1K & 800 && \textbf{41.7}  && \textbf{73.7}  && \textbf{33.8} && \textbf{31.9} \\

\arrayrulecolor{black!15}\midrule[0.25pt]
$\dino$ \citep{caron2021emerging} & $\vit$-B/16 & 85M& IN1K  & 400  && 38.9 && 64.6 && 31.9 && 30.5  \\
$\mae$ \citep{he2022masked} & $\vit$-B/16  & 85M& IN1K  & 1600  && 27.9 && 40.9 && 14.4 && 17.5 \\
\rowcolor{light_cyan} $\mname$ & $\vit$-B/16 & 85M & IN1K  & 400 && \textbf{42.9} && \textbf{74.9}  && \textbf{36.0}  && \textbf{34.7} \\
\arrayrulecolor{black}\midrule[0.5pt]
\end{tabular}
\label{table:linear_segmentation_1token}
}
\end{table}%

%% file: tables/finetune_segmenter.tex
\begin{table}[t]
\centering
\caption{
\textbf{Finetuning evaluation with Segmenter.} Backbones pre-trained with different self-supervised learning methods are finetuned using Segmenter \citep{strudel2021segmenter}. We report the mIoU scores achieved on the validation sets of 4 different datasets. $\dagger$ refers to our own reproduction from official GitHub repositories. If not specified, publicly available checkpoints are used.}

\vspace{0.2cm}
\footnotesize
\resizebox{1\textwidth}{!}{
\begin{tabular}{l c c c c c c c c c c c}

Method & Backbone & Params & Dataset & Epochs && Pascal Context & Pascal VOC & COCO-Stuff 164K & ADE20K \\
\midrule
\textit{Scene-centric} \\
$\dino^{\dagger}$ \citep{caron2021emerging} & $\vit$-S/16 & 21M & COCO & 300 && 33.5 & 66.1 & 35.6 & 35.0 \\
$\mae^{\dagger}$ \citep{he2022masked} & $\vit$-S/16 & 21M & COCO  & 300 && 32.5 & 61.6 & 35.4 & 35.4 \\
$\croc$ \citep{Stegmuller_2023_CVPR} & $\vit$-S/16 & 21M & COCO & 300 && 38.7 & 70.5 & 38.0 & 37.9 \\
\rowcolor{light_cyan} $\mname$ & $\vit$-S/16 & 21M& COCO & 300 && \textbf{41.9} & \textbf{74.2} & \textbf{39.4} & \textbf{39.3} \\
\arrayrulecolor{black!30}\midrule[0.5pt]
\textit{Object-centric} \\
$\dino$ \citep{caron2021emerging} & $\vit$-S/16 & 21M & IN1K & 800 && 46.0 & 80.3 & 43.2 & 43.3 \\
$\croc$ \citep{Stegmuller_2023_CVPR} & $\vit$-S/16 & 21M & IN1K & 300  && 46.0 & 80.9 & 42.9 & 42.8 \\
$\timet$ \citep{salehi2023time} & $\vit$-S/16 & 21M & IN1K+YTVOS  & 800+30 && 47.4 & 80.4 & 43.1 & 43.5 \\
\rowcolor{light_cyan} $\mname$ & $\vit$-S/16 & 21M & IN1K & 800 && \textbf{49.3} & \textbf{82.3} & \textbf{43.9} & \textbf{45.2}  \\
\arrayrulecolor{black!30}\midrule[0.5pt]
$\dino$ \citep{caron2021emerging} & $\vit$-B/16 & 85M & IN1K & 400 && 45.8 & 82.2 &  44.4 & 45.0 \\
$\mae$ \citep{he2022masked} & $\vit$-B/16& 85M & IN1K  & 1600 && 47.9 & 82.7 & \textbf{45.5} & \textbf{46.4} \\
\rowcolor{light_cyan} $\mname$ & $\vit$-B/16 & 85M & IN1K & 400 && \textbf{49.2} & \textbf{83.4} & 44.6 & 46.0 \\
\arrayrulecolor{black}\midrule[0.5pt]
\end{tabular}
\label{table:segmenter_segmentation}
}
\end{table}%

%% file: tables/ablation_hp.tex
\begin{table*}[t]
\centering
\caption{
\textbf{Ablation study on the hyperparameters of $\mname$}. Here, a ViT-S/16 undergoes pretraining on COCO for 300 epochs with various combinations of hyperparameter values. The resulting features are then compared using dense nearest neighbors retrieval. We denote the size of the queues $\mathcal{M}_i$ in terms of the number of images present by $S$. Unless explicitly indicated, the values of the hyperparameters are set to $\lambda_{\text{pos}}=2.0$, $S=25$k, and $K=64$.}
\footnotesize
\resizebox{1\textwidth}{!}{
\begin{tabular}{l c c c c c c c c c c c c c c c c c c}
 & && \multicolumn{6}{c}{Positional weighting ($\lambda_{\text{pos}}$)} && \multicolumn{3}{c}{Memory bank size ($S$)} && \multicolumn{5}{c}{Number of objects ($K$)} \\
\cmidrule{4-9}  \cmidrule{11-13} \cmidrule{15-19}
Dataset & $\mathcal{L}_{ci}^o$ && 0.0 & 0.1 & 1.0 & 2.0 & 4.0 & 8.0 && 1k & 5k & 25k && 4 & 8 & 16 & 32 & 64  \\
\midrule
Pascal VOC & \xk &&  44.9&	46.3&	47.5&	47.5&	45.8&	44.2  && - & - & - && 47.3 & 47.6 & 48.1 & 47.1 & 47.5  \\
\rowcolor{light_cyan} Pascal VOC & \ck &&  55.9&	57.1&	\textbf{58.1}&	\textbf{58.1}&	57.5&	57.0  && 46.7 & 47.3 & \textbf{58.1} && 49.9 & 52.9 & 54.6 & 57.0 & \textbf{58.1}  \\
\arrayrulecolor{black!30}\midrule[0.5pt]
ADE20K & \xk &&  16.9&	18.0&	18.9&	18.4&	17.8&	16.3  && - & - & - && 17.7 & 18.0 & 18.9 & 18.5 & 18.4  \\
\rowcolor{light_cyan} ADE20K & \ck && 21.6&	21.8&	23.1&	\textbf{23.4}&	22.6&	22.5  && 18.1 & 18.4 & \textbf{23.4} && 18.1 & 19.4 & 20.3 & 21.7 & \textbf{23.4}  \\
\arrayrulecolor{black}\midrule[0.5pt]
\end{tabular}
\label{tab:ablation_hp}
}
\end{table*}

%% file: chapters/conclusion.tex
\section{Conclusion}
\label{sec:conclusion}
We introduce CrIBo, the first end-to-end online cross-image object-level bootstrapping self-supervised learning method. CrIBo capitalizes on the augmented sample diversity derived from image-level bootstrapping while harnessing the advantages of localized self-supervision inherent to object-level learning. We extensively evaluate CrIBo across multiple dense downstream tasks showcasing excellent performance on in-context scene understanding tasks and highlighting the merits of the cross-image object-level learning paradigm.

\noindent
{\bf Limitations.}
\label{par:limitations}
This study is developed around the $\vit$ architecture as it has become quite ubiquitous. As such, and for the sake of coherence, we only compare with other $\vit$-based methods. As we have to cope with a computational budget, we chose to invest it in producing a well-grounded and motivated approach with extensive analysis, rather than in exploring larger datasets~\citep{thomee2016yfcc100m,changpinyo2021conceptual} or models, \eg $\vit$-Large or $\vit$-Huge.

%% file: chapters/appendix.tex
\newpage
\appendix

\section*{Appendix}
\section{Experimental setup}
\label{ssec:experimental_setup}

\subsection{Pretraining}
\label{ssec:app_pretraining}

\paragraph{Pre-training datasets.}
\phantomsection
\label{par:pre_training}
Our pretraining datasets include COCO \citep{lin2014microsoft} and ImageNet-1k \citep{5206848}. The former includes about 118k scence-centric images while the latter includes about 1.3M object-centric images. Both are ubiquitous in scene-centric self-supervised learning and object-centric self-supervised learning, respectively.

\paragraph{Network architecture.}
\phantomsection
\label{par:network}
We employ vision transformers as our backbone $f$. This selection aligns with its prevalent use in contemporary methodologies and allows for fair and easy comparisons with existing works. The configuration of the projection heads closely follows that of \cite{caron2021emerging}. Both the image-level head $\overbar{\head}$ and object-level head $\head$ share their weights, except for the final linear layer. Both the image-level and the object-level output distribution have the same size \ie $\overbar{L}=L=65,536$.

\paragraph{Optimization.}
\phantomsection
\label{par:optimization}
The ViT-small (ViT-S/16) is trained for 800 epochs, while the ViT-base (ViT-B/16) is trained for 400 epochs. The ViT-Base is only trained on ImageNet-1k, while the ViT-Small is trained on both COCO and ImageNet-1k. Pretrainings on COCO use a batchsize of 256 while pretrainings on ImageNet-1k use a batchsize of 1024. Learning rate, weight-decay, and other optimization-related hyperparameters are exactly the same as in $\dino$ \citep{caron2021emerging}. 

\paragraph{Hyperparameters.}
\phantomsection
\label{par:hyperparameters}
Results reported in tables using ViT-S/16 (apart from the gridsearch) are based on the following hyperparameters: $(\lambda_{\text{pos}}, S, K)=(1.0, 25\text{k}, 32)$ and $(\lambda_{\text{pos}}, S, K)=(2.0, 25\text{k}, 64)$ for pretrainings on ImageNet-1K and COCO, respectively. Results reported for ViT-B/16 are based using the following hyperparameters: $(\lambda_{\text{pos}}, S, K)=(1.0, 25\text{k}, 32)$.

\subsection{Evaluation protocols}
\label{ssec:app_evaluation_protocols}

\paragraph{Linear segmentation.}
\phantomsection
\label{par:app_linear_segmentation_protocol}
The frozen local representations are fed to the linear decoder head from Segmenter~\citep{strudel2021segmenter}, and we use the implementation available through MMSegmentation~\citep{mmseg2020}. Herein, spatial tokens undergo a linear projection to the class space, followed by bilinear up-sampling to match the input image dimensions, enabling the application of pixel-wise cross-entropy loss. Performance metrics, in terms of mIoU scores, are reported on four different datasets: Pascal Context~\citep{mottaghi2014role}, Pascal VOC 2012~\citep{pascal-voc-2012}, COCO-Stuff 164K~\citep{caesar2018coco} and ADE20K~\citep{zhou2017scene}.
 \par
Regardless of the dataset used, the crop size is always $512\times512$ pixels, and the dataset configurations are kept as provided by MMSegmentation~\citep{mmseg2020}. For ADE20K and COCO-Stuff 164K, we use the \textit{160k\_iterations\_schedule} whereas for Pascal VOC 2012 and Pascal Context we use the \textit{80k\_iterations\_schedule}. The only change we make to the default schedules configurations is that we use the Adam optimizer~\citep{kingma2014adam} instead of SGD. For each pretraining method and dataset, we run with four different learning rates (\texttt{8e-4}, \texttt{3e-4}, \texttt{1e-4}, and \texttt{8e-5}) and report the highest mIoU score.

\paragraph{Finetuning with Segmenter.}
\phantomsection
\label{par:app_finetuning_segmentert}
We perform an end-to-end finetuning of the backbone topped with the transformer-based decoder from Segmenter~\citep{strudel2021segmenter}. Here, the  backbone's spatial features are fed to a transformer encoder along with $K$ learnable class tokens. The resulting class and spatial tokens are projected onto one another to obtain patch-level predictions and subsequently up-sampled to the input image's size to enforce the pixel-wise cross-entropy loss. We report the mIoU scores achieved on 4 different datasets: Pascal Context~\citep{mottaghi2014role}, Pascal VOC 2012~\citep{pascal-voc-2012}, COCO-Stuff 164K~\citep{caesar2018coco} and ADE20K~\citep{zhou2017scene}. \par

Regardless of the dataset used, the crop size is always $512\times512$ pixels, and the dataset configurations are kept as provided by MMSegmentation~\citep{mmseg2020}. For ADE20K and COCO-Stuff 164K, we use the \textit{160k\_iterations\_schedule} whereas for Pascal VOC 2012 and Pascal Context, we use the \textit{80k\_iterations\_schedule}. Two changes are made w.r.t. the default schedules configurations. First, we use the Adam optimizer~\citep{kingma2014adam} instead of SGD and second, we use $\texttt{eta\_min} = 0.1 \cdot \texttt{lr}$ instead of  $\texttt{eta\_min} = \texttt{1e-4}$. For each pretraining method and dataset, we run with four different learning rates (\texttt{8e-5}, \texttt{3e-5}, \texttt{1e-5}, and \texttt{8e-6}) and report the highest mIoU score.

\section{Clustering algorithm}
\label{sec:app_clustering_algorithm}
We briefly detail the clustering algorithm, particularly the distinctions from its original implementation proposed in~\cite{Stegmuller_2023_CVPR}.
 \par
The overall objective of the clustering step is to assign the spatial tokens from \textbf{both} views of a common image to $K$ semantically coherent groups. Formally, given the dense representations $\dense_{1}$ and $\dense_{2}$ of two augmented views, the joint representation $\dense_{\text{cat}}  \in \mathbb{R}^{2N \times d}$ is obtained by a concatenation along the dimension of the tokens. The $k^{th}$ centroid $\cent^{k} \in \mathbb{R}^{d}$ is initialized by uniformly sampling without replacement one of the $2N$ tokens.
The clustering procedure relies on the Sinkhorn-Knopp algorithm~\citep{cuturi2013sinkhorn} to solve the optimal transportation problem of assigning $2N$ tokens to $K$ clusters/centroids. The transportation cost between tokens and centroids $\costsem \in \mathbb{R}^{2N \times K}$ is given by the negative cosine distance:
 \begin{equation}
     \label{eq:cost_new}
     \costsem_{n,k} = - \frac{<\cent^{k}, \dense^{n}_{\text{cat}}>}{\norm{\cent^{k}}\norm{\dense^{n}_{\text{cat}}}}
 \end{equation}
The Sinkhorn-Knopp algorithm efficiently computes a solution (an assignment matrix $\Q^{*}$) to the following optimization problem:
\begin{equation}
    \label{eq:assignments}
    \Q ^{*} = \underset{\Q \in \mathcal{Q}}{\text{argmin}} <\Q, \costsem> - \frac{1}{\lambda} H(\Q)
\end{equation}
which aims to minimize the inner product of the transportation plan and the transportation cost, subject to an entropy constraint on the assignments. The transportation polytope $\mathcal{Q}$ defines the set of valid assignments:
\begin{equation}
    \label{eq:transport_polytope}
    \mathcal{Q}= \{ \Q \in \mathbb{R}^{2N \times K}_{+} \:|\: \Q \mathbf{1}_{K} =  \frac{1}{2N} \mathbf{1}_{2N}, \Q^{\top} \mathbf{1}_{2N} = \frac{1}{K} \mathbf{1}_{K} \}
\end{equation}
We update the centroids by pooling over each cluster:
\begin{equation}
    \label{eq:cent_update}
    \cent^{k} = \sum_{n \in [2N]} \Q^{*}(n,k) \cdot \z_{\text{cat}}^n
\end{equation}
At this point, we could stop, but we observed improved results when doing multiple iterations of the above steps, \ie restarting the procedure from~\Cref{eq:cost_new} and the centroids found in~\Cref{eq:cent_update}.
Empirically, we found that $5$ iterations is a good trade-off between speed and accuracy. In practice, we use the ``hard'' version of the assignments obtained by ensuring that each token is only assigned to a single cluster. After a final $l_{1}$-normalization of the columns of the herein obtained matrix, the view-wise object representations are computed as in~\Cref{eq:vanilla_object_representation}. \par

Importantly, the above-defined algorithm diverges from the one proposed in $\croc$~\citep{Stegmuller_2023_CVPR} in multiple ways. The main difference is that we do not rely on an iterative procedure to find the optimal number of centroids $K$, and as such, the number of clusters is fixed. Another discrepancy is that we use uniform distributions for both the sampling of the initial centroids and the marginals of the assignments (see~\cref{eq:assignments,eq:transport_polytope}). We observed that these modifications simplify the overall algorithm without incurring a drop in performance. In the following paragraph, we discuss our adjustments to the positional cues of the clustering algorithm.

\paragraph{Positional cues}
\label{app_par:position_cues}

An important property of the selected clustering algorithm is its ability to leverage positional cues to compensate for the lack of high-level semantics in the features at the beginning of the training. The spatial guidance is incorporated in the algorithm via an additive constraint on the transportation cost matrix:
\begin{equation}
\costtot = \costsem + \lambda_{\text{pos}} \costpos
\end{equation}
Similar to $\croc$, we use the 2D coordinates $\pos^{(\text{cat})} \in \mathbb{R}^{2N \times 2}$ of every patch/token in the joint dense representation $\dense_{\text{cat}}$ w.r.t. the upper left corner of the image that generated the two views. Up to normalization constants, the distance between the $i^{th}$ token and the $j^{th}$ centroid is given by:
\begin{equation}
    \costpos_{ij} = \underset{l \in \mathcal{S}_{j}}{\argmin} \: || \mathbf{E}^{\text{(cat)}}_{i} -   \mathbf{E}^{\text{(cat)}}_{l}||_{2}
\end{equation}
where $\mathcal{S}_{j}$ is the set of tokens assigned to the $j^{th}$ centroid from the previous iteration. It is noteworthy that the positional cost differs from that of $\croc$, which sets the position of a given cluster as the average position of the tokens assigned to it. In our approach, the distance between a token and a given centroid depends on the query token. This mitigates the issue of concentric objects sharing the same position. \par

In~\Cref{tab:ablation_hp}, we ablate over the hyperparameter $\lambda_{\text{pos}}$, which modulates the importance of the positional cost in the clustering algorithm.
\par

\section{Additional experiments}
\label{sec:app_additional_experiments}

\subsection{Contribution of individual loss terms}
\label{ssec:contribution_loss_term}
\input{tables/ablation_full}

Consecutive evaluations showcase the contribution of each loss in \Cref{tab:ablation_loss} on 3 different downstream tasks. It follows that all losses contribute a significant improvement to the overall performance of CrIBo.

\subsection{Object detection with YOLO-S}
\label{ssec:app_object_detection_yolos}
\input{tables/object_detection_yolos}
We consider an object detection downstream task to shed a different light on the quality of the learned embeddings. We perform an end-to-end finetuning of the backbone using YOLO-S~\citep{fang2021you} on COCO~\cite{lin2014microsoft}. We assess performance based on the standard Average Precision metrics, namely $\ie$ $\text{AP}^{\text{bb}}_{\text{50}}$, $\text{AP}^{\text{bb}}_{\text{75}}$ and $\text{AP}^{\text{bb}}$, computed from bounding boxes predicted on the validation set. We strictly follow the finetuning protocol from \cite{fang2021you}, with the sole modification being an adjustment to a batchsize of 32, as opposed to the originally used batchsize of 8. 
 \par 

Our choice of object detection method was guided by its simplicity and minimal requirement for additional parameters, aiming to capture the nuances of various pretraining methods. However, \Cref{table:yolos} reveals that YOLO-S~\citep{fang2021you} is relatively insensitive to the pretraining scheme and provides less information than anticipated. In hindsight, it might have been more judicious to start with the parameters specific to YOLO-S already trained. Indeed, this approach involves appending a hundred detection tokens to the image tokens before transmitting the resulting sequence through the transformer encoder. As such, earlier layers are at risk of receiving strong and biased gradients at the beginning of the finetuning phase, potentially overriding the properties learned during pretraining and explaining the observed results. On the other hand, the results in~\Cref{table:yolos} are typically in the ballpark of those reported in the original paper~\citep{fang2021you}.

\input{tables/frozen_linear_segmentation_4tokens}

\input{tables/ablation_bt}
\subsection{Segmentation with a linear head and concatenation}
\label{sssec:app_linear_segmentation_eval}
We replicate the same linear segmentation experiment from the main text (see~\Cref{sssec:linear_segmentation_eval}), but we concatenate the spatial tokens from the last 4 layers of the $\vit$s in \Cref{table:linear_segmentation_4tokens}. This scenario yields improved overall performance, but it's not clear whether these improvements result from the fact that different layers of the $\vit$ encode different information or from the increased number of trainable parameters.

\begin{figure}[t]
    \centering
    \includegraphics[width=\textwidth]{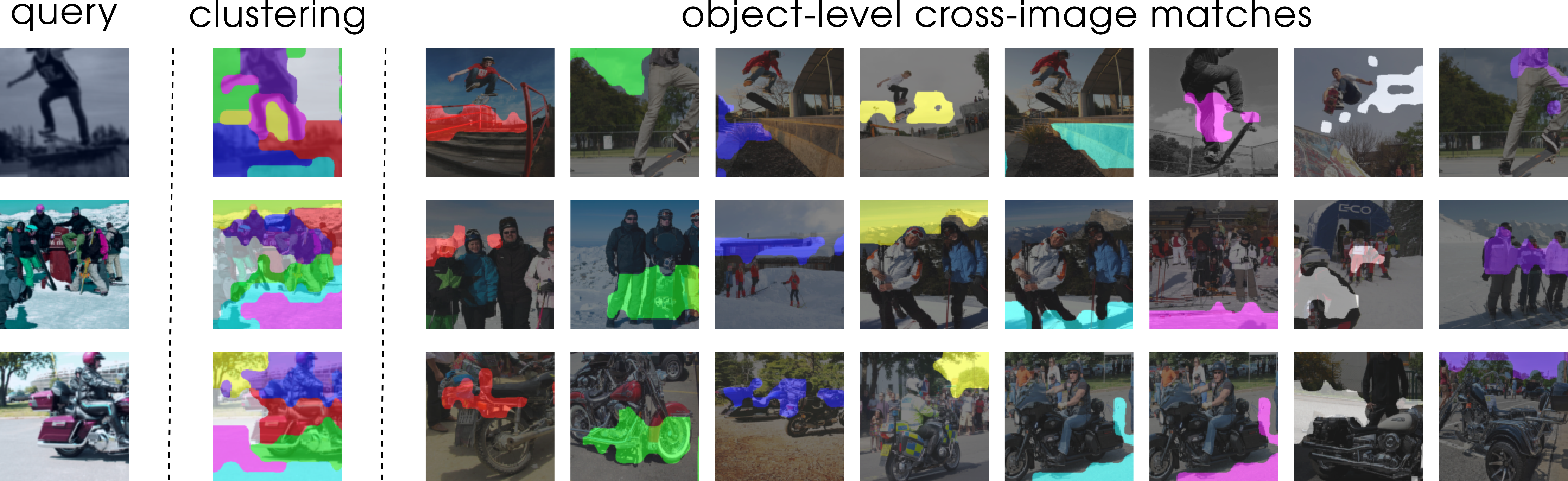}
    \caption{
    \textbf{Visualization of cross-image object-level matchings on COCO.} For a given query view (considered as view 1 here), object representations are computed via clustering. For each object-representation $\cent_1^k$ (highlighted in unique colors), its nearest neighbor $\NN(\cent_1^k, \mathcal{M}_1)$ in the memory bank $\mathcal{M}_1$ is visualized. In this visualization, $K=12$.
    }%
    \label{fig:matching_coco}
\end{figure}

\begin{figure}[t]
    \centering
    \includegraphics[width=\textwidth]{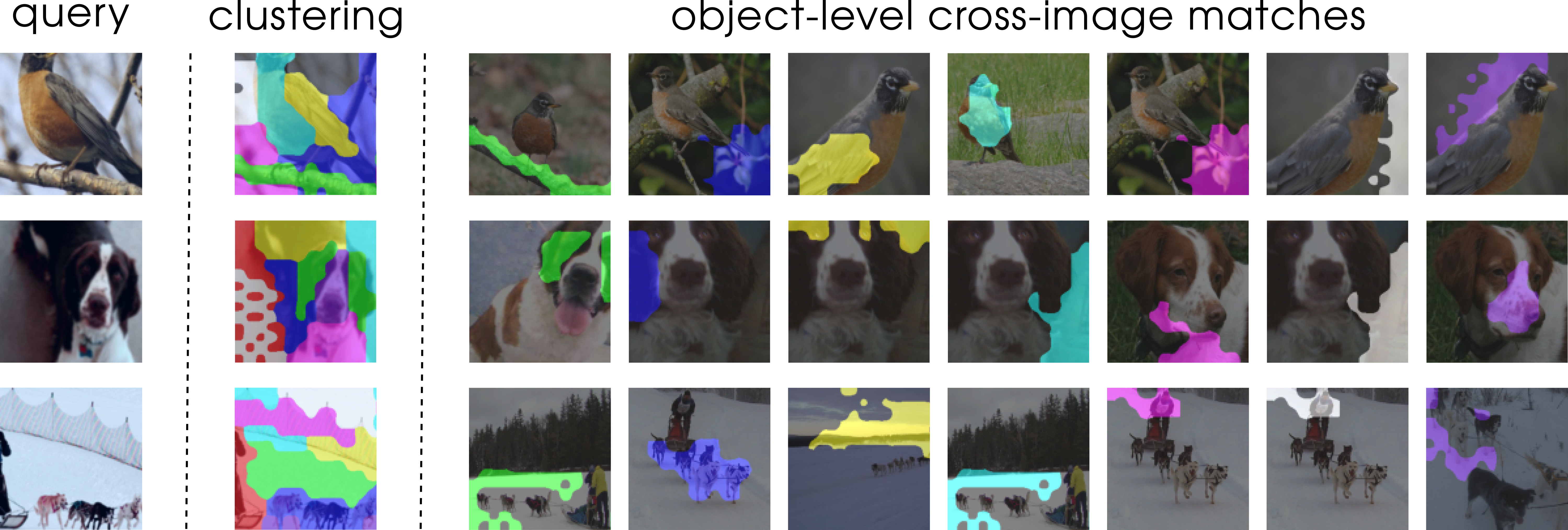}
    \caption{
    \textbf{Visualization of cross-image object-level matchings on ImageNet-1k.} For a given a query view (considered as view 1 here), object representations are computed via clustering. For each object-representation $\cent_1^k$ (highlighted in unique colors), its nearest neighbor $\NN(\cent_1^k, \mathcal{M}_1)$ in the memory bank $\mathcal{M}_1$ is visualized. In this visualization, $K=12$.}
    \label{fig:matching_in}
\end{figure}

\label{sec:app_additional_ablations}
\subsection{Cycle consistency brings stability}
\label{ssec:ablationstudy_cc}
We evaluate the effect of the cycle consistency criterion by comparing the results of various pretraining settings with and without the bootstrapping condition. Training is performed on ImageNet-1k for 300 epochs for every combination of the following hyperparameters:
\begin{itemize}
    \item[-] $K$: $\{4, 8, 12\}$.
    \item[-] $\lambda_{\text{pos}}$: $\{1., 2., 4. \}$.
    \item[-] $S$: $\{25\text{k}, 50\text{k}\}$.
\end{itemize}
We then evaluate the resulting models with the dense nearest neighbor retrieval task (see~\Cref{sssec:dense_nn_eval}). In~\Cref{tab:ablation_bt}, we report the aggregated mIoU results for two datasets (Pascal VOC 2012~\citep{pascal-voc-2012} and ADE20K~\citep{zhou2017scene}) and the two settings, \ie with and without the cycle consistency condition. The aggregation is either the minimum, the maximum, or the average of all the results. As can be observed in~\Cref{tab:ablation_bt}, the proposed criterion does not bring improvements \textit{per se}, but stabilizes the training procedure. Indeed, under the systematic bootstrapping regime, the performance can be unpredictable, as exemplified by the large variations of mIoU scores. This happens when nearest neighbors are determined based on low-level cues, a scenario that adaptive bootstrapping strives to prevent through the employment of cycle consistency.

\subsection{Visualization of cross-image object-level matchings}
\label{ssec:imagenetmatchings}
\Cref{fig:matching_coco} and \Cref{fig:matching_in} show the cross-image object-level matchings during a pretraining on COCO and ImageNet-1k, respectively. Even though ImageNet-1k is an object-centric dataset, classes can always be decomposed into finer grained classes. For example, the first image with the bird has tree branches present, which CrIBo does not fail to recognize.

\begin{figure}[t]
    \centering
    \includegraphics[width=\textwidth]{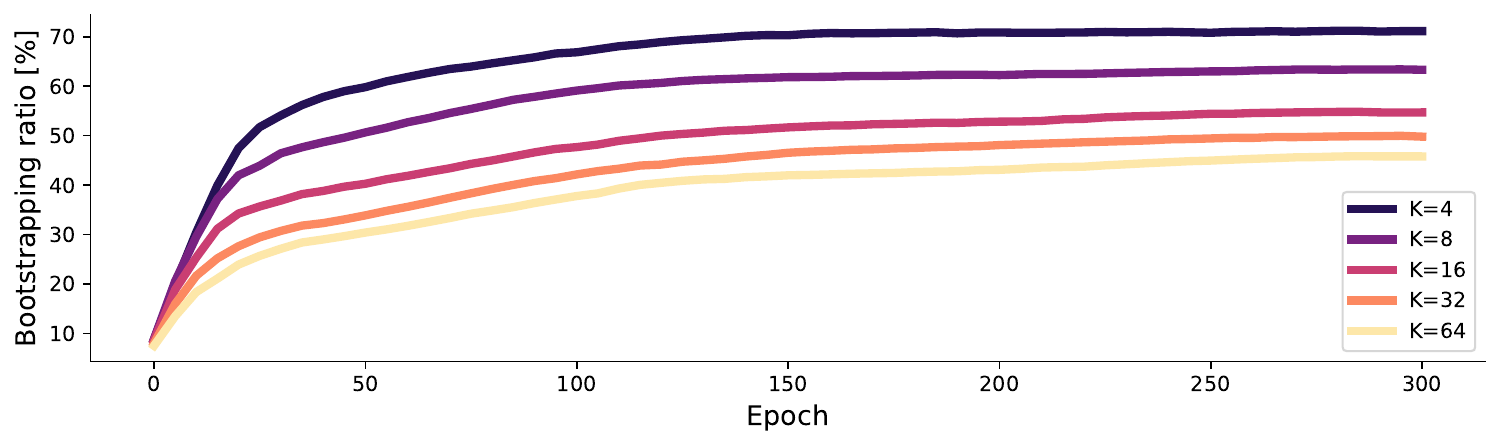}
    \caption{\textbf{Bootstrapping ratio over the epochs.} More bootstrapping is allowed as the training progresses, and as the representations become more semantically driven.}
    \label{fig:bootstrapping_ratio}
\end{figure}
\input{tables/profiling}

\subsection{Bootstrapping ratio}
\label{ssec:bootstrapping_ratio}
The evolution of the bootstrapping ratio over the training epochs is illustrated in \Cref{fig:bootstrapping_ratio}. Using the notations from the paper, the bootstrapping ratio corresponds to approximately $\frac{Z_1+Z_2}{2K}$ averaged over the whole dataset. $Z_1$ and $Z_2$ are the normalization constants from \Cref{eq:dense_cross_image_loss}. In practice, the number of object representations in an image is slightly smaller than $K$ as we discard the ones that do not span both views. At the initial epoch, the encoders are randomly initialized, making it very hard for the object representations to satisfy the cycle consistency condition. However, as learning progresses, representations become more semantically driven, and more bootstrapping is allowed. It is worth noting that a higher $K$ results in a lower bootstrapping ratio as both the batch and the queue are larger (in terms of object representations), which imposes a more stringent condition to meet.

\subsection{High-level profiling}
A detailed timing analysis of CrIBo is performed in \Cref{table:time_oh}. It is worth noting that CriBo-specific operations take less than 15\% of the total time.

\subsection{Runtime comparison}
\label{ssec:runtime_comparison}
\input{tables/runtime_comparison}
A runtime comparison with other methods can be found in \Cref{table:runtime_comparison}. CrIBo is faster than DINO and slower than MAE. Note that the quantitative comparisons in our paper are done with MAE trained for 1600 epochs and DINO/CrIBo trained for 400 epochs. This leads to a similar total computational budget.

\subsection{Supervised oracles}
\label{ssec:overclustering}
\input{tables/ablation_clustering}

An ablation using different supervised oracles is shown in \Cref{tab:overclustering}. The supervision takes place at two different levels: 1) the clustering and 2) the bootstrapping criterion.
\begin{enumerate}
    \item Instead of using an unsupervised clustering algorithm for computing object representations, the semantic segmentation labels from COCO and COCO-Stuff are used. The masks are downsampled (16x) to match the dense representation output resolution. Given an image, the object representation associated with a given mask label is the average pooling of the patches assigned to the label.
    \item Instead of using the cycle-consistency condition for discarding supposed invalid pairs of NNs, the downsampled (16x) semantic masks of COCO and COCO-Stuff are used. A pair of NN object representations is deemed valid if the union of all patch labels is the same in both object representations.
\end{enumerate}
Surprisingly, using unsupervised clustering results in better performance than using the labels from COCO. In hindsight, this confirms again that the overclustering regime is where CrIBo is most at ease. Indeed, the granularity of the mask annotations in COCO is not particularly high. The results would have probably been different using finer-grained annotations. However, replacing the cycle-consistency condition with a supervised oracle yields modest improvements.

\subsection{Clustering algorithm}
\label{ssec:clustering_algorithm}
\input{tables/clustering_ablation}

The clustering algorithm plays a pivotal role in the overall performance of $\mname$. We compare the clustering algorithm, detailed in~\Cref{sec:app_clustering_algorithm}, with a well-established alternative, \ie the K-Means algorithm. Consequently, we pretrain $\vit$-S/16 for $300$ epochs on COCO~\citep{lin2014microsoft}, using a memory bank of size $S=25$k, various numbers of clusters $K$ along with the K-Means algorithm.
 \par

\Cref{tab:clustering_ablation} depicts the performance achieved by each clustering algorithm on the dense nearest neighbor retrieval task (see~\cref{sssec:dense_nn_eval}). It can be observed that using the Sinkhorn-based clustering from \Cref{sec:app_clustering_algorithm} leads to an improvement over the K-Means algorithm.
This difference is only partially explained by the lack of positional cues in K-Means as shown by the row $\lambda_{\text{pos}} = 0.0$.
Indeed, the Sinkhorn-based method, deprived of positional cues, still yields better scores, confirming the utility of the non-standard clustering algorithm.

\section{Datasets}
\label{sec:app_datasets}

\paragraph{Pascal VOC 2012~\citep{pascal-voc-2012}}
\label{par:app_pascal_voc_2012}
A dataset composed of a training set includes 10,582 images distributed across 21 classes, with one being a background class. The validation set incorporates 1,449 images.

\paragraph{Pascal Context~\citep{mottaghi2014role}}
\label{par:app_pascal_context}
A scene-centric dataset comprises 4,998 training images spanning 60 semantic classes (including the background). The validation set contains 5,105 images.

\paragraph{COCO-Stuff 164K~\citep{caesar2018coco}}
\label{par:app_coco_stuff164k}
A scene-understanding dataset, featuring labels spanning 91 ``stuff'' categories and 80 ``things'' categories. The training set is composed of 118K images, while the validation set contains 5K images.

\paragraph{ADE20K~\citep{zhou2017scene}}
\label{par:app_ade20k}
A dataset that encompasses scenes featuring fine-grained labels across 150 distinct semantic categories, making it one of the most demanding available semantic segmentation datasets. The training set comprises 20,210 images, and the validation set consists of 2,000 images.

%% file: tables/ablation_full.tex
\begin{table*}[t]
\centering
\caption{\textbf{Ablation study on the individual contribution of each loss.} $\vit$-S/16 are trained on COCO for 300 epochs and evaluated with downstream tasks of different levels of granularity.}
\footnotesize
\resizebox{1\textwidth}{!}{
\begin{tabular}{c c c c c c c c c c c}
 & & & \multicolumn{2}{c}{Patch-level NN} && \multicolumn{3}{c}{Object Detection with YOLO-S} && \multicolumn{1}{c}{Image-level NN} \\
 \cmidrule{4-5}  \cmidrule{7-9} \cmidrule{11-11}
$\mathcal{L}_{cv}^{g}$ & $\mathcal{L}_{cv}^{o}$ & $\mathcal{L}_{ci}^{o}$ & Pascal VOC (mIoU) & ADE20k (mIoU) && COCO ($\text{AP}^{\text{bb}}$) & COCO ($\text{AP}^{\text{bb}}_{\text{50}}$) & COCO ($\text{AP}^{\text{bb}}_{\text{75}}$) && IN1K (top-1 Acc.) \\
 \midrule
\ck & \xk & \xk & 31.6 & 12.8 && 24.0 & 40.6 & 23.9 && 34.3  \\
\ck & \ck & \xk & 47.1 & 18.3 && 27.1 & 44.2 & 27.6 && 35.8 \\
\rowcolor{light_cyan}\ck & \ck & \ck & \textbf{58.1} & \textbf{23.4} && \textbf{30.2} & \textbf{48.4} & \textbf{31.0} && \textbf{38.2} \\
\arrayrulecolor{black}\midrule[0.5pt]
\end{tabular}
\label{tab:ablation_loss}
}
\end{table*}

%% file: tables/object_detection_yolos.tex
\begin{table}[t]
\centering
\caption{
\textbf{Object detection with YOLO-S.} Backbones pre-trained with different self-supervised learning methods are finetuned with YOLO-S~\citep{he2017mask}. We report the mean average precision for the predicted bounding boxes ($\text{AP}^{\text{bb}}$) for different IoU thresholds on COCO.
}
\vspace{0.2cm}
\footnotesize
\resizebox{1\textwidth}{!}{
\begin{tabular}{l c c c c c c c c c}
Method & Backbone & Params & Dataset & Epochs && $\text{AP}^{\text{bb}}$ & $\text{AP}^{\text{bb}}_{\text{50}}$ & $\text{AP}^{\text{bb}}_{\text{75}}$  \\
\midrule
\textit{Object-centric} \\
$\dino$ \cite{caron2021emerging} & $\vit$-S/16 & 21M & IN1K & 800 && 35.4 & \textbf{56.0} & 36.6 \\
$\croc$ \cite{Stegmuller_2023_CVPR} & $\vit$-S/16 & 21M & IN1K & 300 && \textbf{35.7} & 55.5 & 37.0 \\
$\timet$ \cite{salehi2023time} & $\vit$-S/16 & 21M & IN1K+YTVOS & 800+30 && 35.0 & 55.5 & 36.2 \\
\arrayrulecolor{black!30}\midrule[0.5pt]
\textit{Ours} \\
\rowcolor{light_cyan} $\mname$ & $\vit$-S/16 & 21M & IN1K & 800 && \textbf{35.7} & 55.6 & \textbf{37.4} \\
\arrayrulecolor{black}\midrule[0.5pt]
\end{tabular}
\label{table:yolos}
}
\end{table}%

%% file: tables/frozen_linear_segmentation_4tokens.tex
\begin{table}[b]
\centering
\caption{
\textbf{Linear segmentation with frozen backbones.} The linear decoder from Segmenter~\citep{strudel2021segmenter} is trained on the frozen and concatenated spatial features from the last 4 layers of $\vit$s pretrained with various self-supervised learning methods. We report the mIoU scores achieved on the validation sets of 4 different datasets. Publicly available checkpoints are used when possible, and our reproduced results are indicated with a $\dagger$ symbol.
}
\vspace{0.2cm}
\footnotesize
\resizebox{1\textwidth}{!}{
\begin{tabular}{l c c c c c c c c c c c c c c c}
Method & Backbone & Params & Epochs & Dataset && Pascal Context && Pascal VOC && COCO-Stuff 164K && ADE20K \\
\midrule
\textit{Scene-centric} \\
$\dino^{\dagger}$ \citep{caron2021emerging} & $\vit$-S/16 & 21M & 300 & COCO && 28.9 && 47.4 &&  21.6 && 20.3 \\
$\mae^{\dagger}$ \citep{he2022masked} & $\vit$-S/16 & 21M & 300 & COCO && 19.9 && 29.4 && 13.0 && 13.5 \\
$\croc$ \citep{Stegmuller_2023_CVPR} & $\vit$-S/16 & 21M & 300 & COCO &&  32.2 && 54.7 && 25.9 &&  24.6  \\
\rowcolor{light_cyan} $\mname$ & $\vit$-S/16 & 21M & 300 & COCO && \textbf{37.9} && \textbf{62.4} && \textbf{30.3} && \textbf{29.9}  \\
\arrayrulecolor{black!30}\midrule[0.5pt]
\textit{Object-centric} \\
$\dino$ \citep{caron2021emerging} & $\vit$-S/16 & 21M & 800 & IN1K &&  41.3 &&  69.0 && 33.0 &&  31.0 \\
$\croc$ \citep{Stegmuller_2023_CVPR} & $\vit$-S/16 & 21M & 300 & IN1K  && 39.9 && 71.4 && 33.4 && 31.3  \\
$\timet$ \citep{salehi2023time} & $\vit$-S/16 & 21M & 800+30 & IN1K+YTVOS && 41.9 && 71.9 && 34.5 && 32.3  \\
\rowcolor{light_cyan} $\mname$ & $\vit$-S/16 & 21M & 800 & IN1K && \textbf{42.4} && \textbf{75.2} && \textbf{34.6} && \textbf{33.4}  \\
\arrayrulecolor{black!15}\midrule[0.25pt]
$\dino$ \citep{caron2021emerging} & $\vit$-B/16 & 85M & 400 & IN1K && \textbf{43.1} && 74.2 && 29.9 &&  34.5 \\
$\mae$ \citep{he2022masked} & $\vit$-B/16 & 85M & 1600 & IN1K  && 32.7 &&  53.0 && 19.7 && 23.5  \\
\rowcolor{light_cyan} $\mname$ & $\vit$-B/16 & 85M & 400 & IN1K && 43.0 && \textbf{75.7} && \textbf{36.5} && \textbf{35.4} \\
\arrayrulecolor{black}\midrule[0.5pt]
\end{tabular}
\label{table:linear_segmentation_4tokens}
}
\end{table}%

%% file: tables/ablation_bt.tex
\begin{table*}[b]
\centering
\caption{\textbf{Ablation study on the adaptive bootstrapping using the cycle consistency condition.} Results are aggregated over multiple hyperparameters. Trained on ImageNet-1k for 300 epochs and evaluated using dense nearest neighbor retrieval (\cref{table:hummingbird}).}
\footnotesize
\resizebox{1\textwidth}{!}{
\begin{tabular}{l c c c c c c c c }
 && \multicolumn{3}{c}{Pascal VOC} && \multicolumn{3}{c}{ADE20k} \\
 \cmidrule{3-5}  \cmidrule{7-9}
Bootstrap criterion && min. mIoU & max. mIoU & avg. mIoU && min. mIoU & max. mIoU & avg. mIoU  \\
\midrule
None && 49.1 & 71.2 & 68.8 && 16.4 & 27.2 & 25.7 \\
\rowcolor{light_cyan} Cycle-consistency && \textbf{68.0} & \textbf{71.5} & \textbf{69.9} && \textbf{24.8} & \textbf{27.6} & \textbf{26.0}  \\
\arrayrulecolor{black}\midrule[0.5pt]
\end{tabular}
\label{tab:ablation_bt}
}
\end{table*}%

%% file: tables/profiling.tex
\begin{table}[t]
\centering
\caption{\textbf{High-level profiling of CrIBo}. Experiments are run on a single node with 4x AMD MI250x (2 compute die per GPU \ie $\texttt{worldsize}=8$) with a memory usage of 43.5 GB per compute die. The backbone is a ViT-B/16 and the batchsize is set to 128 per compute die \ie 1024 in total and $K=12$. CrIBo specific operations are \hl{highlighted} (operations specific to object-level nearest neighbors).
}
\footnotesize
\begin{tabular}{l l c c c}
Description & Operation & Absolute time per iteration [ms] && Relative time [\%] \\
\midrule
Forward pass &$f_{t}(\cdot) + f_{s}(\cdot) + \overbar{h}_{t}(\cdot) + \overbar{h}_{s}(\cdot)$  & 453.4 && 41.7 \\
\rowcolor{cyan!20} Clustering & $\Q^{*} = \mathcal{C}(\cdot)$ & 82.2 && 7.6\\
\rowcolor{cyan!20} NN retrieval & $\texttt{nn}(\cdot)$ + $\texttt{update}(\mathcal{M}_i)$ & 47.5 && 4.4 \\
\rowcolor{cyan!20} Object projections & $h_t(\cdot)$ + $h_s(\cdot)$ & 11.2 && 1.0 \\
Weights update & \texttt{backprop.} + \texttt{EMA} & 493.1 && 45.3 \\
\arrayrulecolor{black!30}\midrule[0.5pt]
\textit{total} & & 1087.4 & & 100\\
\arrayrulecolor{black}\midrule[0.5pt]
\end{tabular}
\label{table:time_oh}
\end{table}

%% file: tables/runtime_comparison.tex
\begin{table}[b]
\centering
\caption{\textbf{Runtime comparison.} Experiments are run on a single node with 4x AMD MI250x (2 compute die per GPU i.e. $\texttt{worldsize}=8$).}
\footnotesize
\begin{tabular}{l c c c}
Method & Time per epoch [minutes:seconds] & Memory per compute die [GB] & Batchsize \\
\midrule
MAE & 04:45 & $\sim36$ & 4096 \\
DINO & 29:43	 & $\sim45$ & 1024 \\
CrIBo ($K=12$) & 23:58	 & $\sim44$ & 1024 \\
\rowcolor{cyan!20} CrIBo ($K=32$) & 28:22	 & $\sim60$ & 1024 \\
\arrayrulecolor{black}\midrule[0.5pt]
\end{tabular}
\label{table:runtime_comparison}
\end{table}

%% file: tables/ablation_clustering.tex
\begin{table*}[t]
\centering
\caption{
\textbf{We investigate the usage of supervision in $\mname$.} To get a sense of how supervision could help CrIBo, we ablate over the usage of labels for the clustering step and for the bootstrapping criterion. The hyperparameters (where applicable) are set to $(\lambda_{\text{pos}}, S, K)=(2.0, 25\text{k}, 64)$.}
\footnotesize
\begin{tabular}{ c c c c c}
\multicolumn{2}{c}{Supervision} && \multicolumn{2}{c}{Patch-level NN}  \\
 \cmidrule{1-2}  \cmidrule{4-5} 
  Clustering &  Bootstrapping && Pascal VOC & ADE20k  \\
 \midrule
  \ck    & \ck && 52.5 & 21.2 \\
  \ck    & \xk && 52.3 & 20.2 \\
  \xk    & \ck && \textbf{59.5} & \textbf{23.4} \\
 \rowcolor{light_cyan}   \xk  & \xk && 58.1 & \textbf{23.4} \\
\arrayrulecolor{black}\midrule[0.5pt]
\end{tabular}
\label{tab:overclustering}
\end{table*}

%% file: tables/clustering_ablation.tex
\begin{table*}[b]
\centering
\caption{
Ablation study on the clustering algorithm.
}
\footnotesize
\resizebox{1\textwidth}{!}{
\begin{tabular}{l c c c c c c c c c c c c c c c}
 &&  \multicolumn{2}{c}{$K=4$} && \multicolumn{2}{c}{$K=8$} && \multicolumn{2}{c}{$K=16$} && \multicolumn{2}{c}{$K=32$}  && \multicolumn{2}{c}{$K=64$}\\
\cmidrule{3-4}  \cmidrule{6-7} \cmidrule{9-10} \cmidrule{12-13} \cmidrule{15-16}
Clustering && Pascal VOC & ADE20K && Pascal VOC & ADE20K && Pascal VOC & ADE20K && Pascal VOC & ADE20K && Pascal VOC & ADE20K \\
\midrule
 K-Means && 42.8 & 12.5 && 47.6 & 13.8 && 49.1 & 14.3 && 52.0 & 15.3 && 53.0 & 16.0 \\
\arrayrulecolor{black!30}\midrule[0.5pt]
Sinkhorn ($\lambda_{\text{pos}} = 0.0$) && 47.2 & 16.7 && 50.1 & 17.7 && 52.4 & 18.9 && 53.8 & 20.1 && 55.9 & 21.6 \\
\rowcolor{light_cyan} Sinkhorn ($\lambda_{\text{pos}} = 2.0$) && \textbf{49.9} & \textbf{18.1} && \textbf{52.9} & \textbf{19.4} && \textbf{54.6} & \textbf{20.3} && \textbf{57.0} & \textbf{21.7} && \textbf{58.1} & \textbf{23.4} \\
\arrayrulecolor{black}\midrule[0.5pt]
\end{tabular}
\label{tab:clustering_ablation}
}
\end{table*}%